\documentclass[letterpaper, 10 pt, journal, twoside]{IEEEtran}

\usepackage[table,usenames,dvipsnames]{xcolor}
\usepackage{enumitem}
\usepackage[noadjust]{cite}
\usepackage{amsmath,amssymb,amsfonts,amsthm,amscd,dsfont} 
\usepackage[makeroom]{cancel}
\usepackage{mathtools}

\usepackage{algorithm,algorithmicx,listings}        
\usepackage[noend]{algpseudocode}			        

\usepackage{graphicx,tabularx,adjustbox}
\graphicspath{{./fig/}}
\usepackage{multicol}
\usepackage{multirow}
\usepackage{makecell}
\usepackage{booktabs}
\usepackage{siunitx}
\usepackage[font={small}]{caption}   
\usepackage{subcaption}

\usepackage{tikz}
\usetikzlibrary{shapes,arrows}

\usepackage{afterpage}
\usepackage{balance}

\usepackage[bookmarks=true, breaklinks=true, colorlinks]{hyperref}

\newtheorem*{proposition*}{Proposition}

\newtheorem*{corollary*}{Corollary}

\theoremstyle{definition}

\newtheorem*{assumption*}{Assumption}
\newtheorem{problem}{Problem}
\newtheorem*{problem*}{Problem}
\theoremstyle{remark}

\newcommand{\calD}{{\cal D}}

\newcommand{\calP}{{\cal P}}

\newcommand{\calX}{{\cal X}}

\newcommand{\bbR}{\mathbb{R}}

\newcommand{\NEW}[1]{{\color{black}#1}}

\begin{document}
%
\title{Learned IMU Bias Prediction for Invariant Visual Inertial Odometry}

\author{Abdullah Altawaitan$^1$, Jason Stanley$^1$, Sambaran Ghosal$^1$, Thai Duong$^2$, and Nikolay Atanasov$^1$%
\thanks{This work has been submitted to the IEEE for possible publication. Copyright may be transferred without notice, after which this version may no longer be available.}
\thanks{We gratefully acknowledge support from NSF CCF-2112665 (TILOS).}
\thanks{$^1$The authors are with the Department of Electrical and Computer Engineering, University of California San Diego, La Jolla, CA 92093, USA, e-mails: {\tt\small \{aaltawaitan,\allowbreak jtstanle,\allowbreak sghosal,\allowbreak natanasov\}@ucsd.edu}. A. Altawaitan is also affiliated with Kuwait University as a holder of a scholarship.}
\thanks{$^2$This author is with the Department of Computer Science, Rice University, Houston, TX 77005, USA, e-mail: {\tt\small thaiduong@rice.edu}.}
}

\maketitle

\begin{abstract}
Autonomous mobile robots operating in novel environments depend critically on accurate state estimation, often utilizing visual and inertial measurements. Recent work has shown that an invariant formulation of the extended Kalman filter improves the convergence and robustness of visual-inertial odometry by utilizing the Lie group structure of a robot's position, velocity, and orientation states. However, inertial sensors also require measurement bias estimation, yet introducing the bias in the filter state breaks the Lie group symmetry. In this paper, we design a neural network to predict the bias of an inertial measurement unit (IMU) from a sequence of previous IMU measurements. This allows us to use an invariant filter for visual inertial odometry, relying on the learned bias prediction rather than introducing the bias in the filter state. We demonstrate that an invariant multi-state constraint Kalman filter (MSCKF) with learned bias predictions achieves robust visual-inertial odometry in real experiments, even when visual information is unavailable for extended periods and the system needs to rely solely on IMU measurements.
\end{abstract}


\IEEEpeerreviewmaketitle


\section{Introduction}
\IEEEPARstart{M}{any} core robot autonomy functions, including mapping and control, depend on accurate state estimation. Visual-inertial odometry (VIO) \cite{delmerico2018vio,huang2019vio} offers a reliable and cost-effective approach to estimate the position, orientation, and velocity of mobile robots equipped with cameras and inertial measurement units (IMUs). Cameras can estimate pose displacements but are sensitive to lighting change and motion blur. IMUs, on the other hand, deliver high-frequency data independent of visual conditions but lead to estimate drift over time due to measurement bias. Thus, visual and inertial sensors complement each other effectively but estimating IMU bias is crucial for ensuring reliable state estimation, especially with poor or intermittent visual information.

Traditional VIO methods like the multi-state constraint Kalman filter (MSCKF) \cite{mourikis2007multi} include the IMU bias, together with the system position, orientation, and velocity, in the filter state and estimate it sequentially from sensor measurements. We explore an alternative formulation using a learned sequence-to-sequence model to predict the IMU bias based on a longer history of IMU measurements. Moreover, VIO systems typically model IMU bias as a random process driven by white noise rather than as an unknown term. This distinction impacts the observability properties of the VIO system: bias is observable when modeled as noise but unobservable when considered as an unknown term, as shown in \cite{hernandez2015observability}. In practice, IMU biases often exhibit slow time-varying drift rather than the rapid fluctuations characteristic of white noise. In this work, we design a neural network that predicts IMU biases directly from a sequence of previous IMU measurements, avoiding the need to include the bias in the filter state and allowing the use of an invariant Kalman filter formulation as we discuss later. 

\begin{figure}[t]
    \centering
    \includegraphics[width=1.0\linewidth]{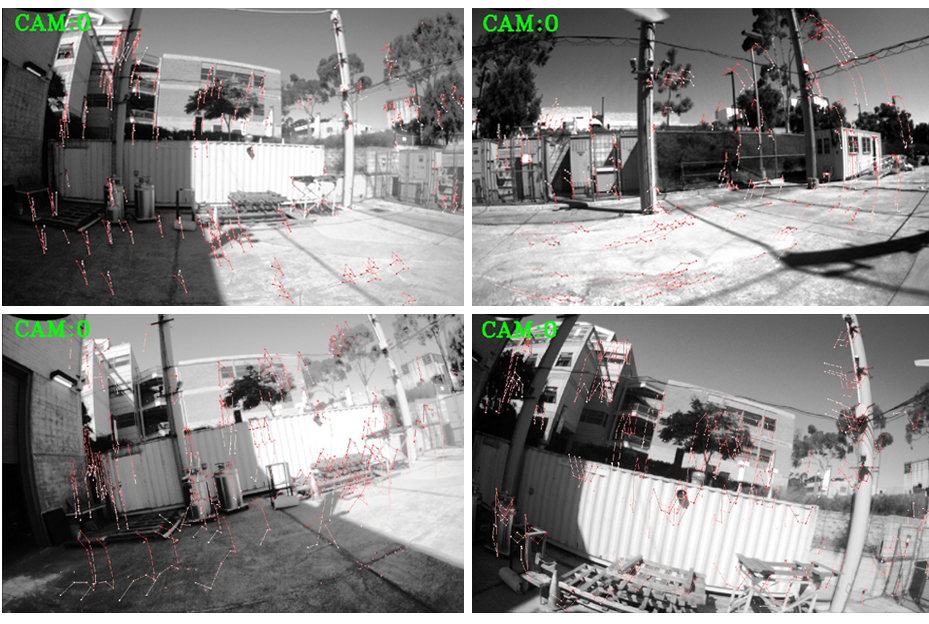}
    \caption{Monocular images and keypoints from a quadrotor with a FLIR Chameleon camera and VectorNav VN-100 IMU. 
    }
    \label{fig:thumbnail}
    \vspace{-5mm}
\end{figure}

First, we review learning-based methods that leverage IMU data for state estimation. IONet \cite{chen2018ionet} uses a long-short-term memory (LSTM) network to predict velocities from buffered IMU measurements that are then integrated to estimate the 2D motion of pedestrians. RoNIN \cite{herath2020ronin} continues this direction by presenting three different neural network architectures: Temporal Convolutional Network (TCN), Residual Network (ResNet), and LSTM to predict velocities which, when integrated with known orientation, yield 2D pedestrian motion estimates. TLIO \cite{liu2020tlio} extends previous works to 3D and uses a ResNet to estimate pedestrian displacements in a local gravity-aligned frame and their uncertainty from a buffer of IMU measurements, which serve as measurements in an extended Kalman filter (EKF). While \cite{chen2018ionet, herath2020ronin, liu2020tlio} focus on pedestrian motions, Zhang et al. \cite{zhang2022dido} show that a series of neural networks can be used to estimate IMU bias, thrust correction, and integration errors for a quadrotor robot using IMU readings and motor speeds. Likewise, Cioffi et al. \cite{cioffi2023learned} use a TCN network to predict 3D relative position from thrust and gyroscope measurements for drone racing. However, the methods in \cite{chen2018ionet, herath2020ronin, liu2020tlio, zhang2022dido, cioffi2023learned} are trajectory-specific and cannot generalize to unseen trajectories at test time. Moreover, these methods often assume that the IMU measurements are transformed into the world frame using the ground-truth pose but, during deployment, the pose is typically estimated via Kalman filtering, making the transformation inaccurate. To address this limitation, Buchanan et al. \cite{buchanan2022deep} propose a neural network to predict the IMU bias directly instead of learning a motion model, enabling the system to generalize to unseen trajectories at test time. However, the network is trained with ground-truth IMU biases, which are unavailable in real-world scenarios. Qiu et al. \cite{qiu2023airimu} extend \cite{buchanan2022deep} by learning both IMU bias and measurement uncertainty through IMU preintegration in pose graph optimization. Denoising IMU Gyroscopes \cite{brossard2020denoising} learns only the gyroscope bias, which is not enough for accurate inertial integration, and evaluates rotational accuracy alone. In contrast, we learn both gyroscope and accelerometer biases and evaluate both translational and rotational accuracy. TLIO \cite{liu2020tlio} trains a network to estimate relative positions directly from IMU measurements in a local gravity-aligned frame, implicitly learning the unobserved initial velocity in a time window from pedestrian motion patterns. Instead, we train a neural network to estimate IMU bias from past IMU measurements using a Lie algebra error between the integrated measurements and the ground-truth robot state. DIDO \cite{zhang2022dido} learns biases separately for gyroscope and accelerometer and relies on a tachometer, prior knowledge of quadrotor parameters, and near hover flight assumption, and decouples rotational and translational dynamics to estimate relative positions and velocities for inertial-only odometry. DIDO overfits to motion patterns seen during training due to the unobserved initial velocity, similar to TLIO. Our method removes these requirements by predicting both biases with a single bias prediction network and integrating it into visual-inertial odometry. Finally, AirIMU \cite{qiu2023airimu} trains a network to estimate the IMU bias for an IMU-GPS pose graph optimization. However, AirIMU assumes access to ground-truth positions during deployment and does not claim real-time performance. In contrast, we learn the bias prediction model in the same way but integrate it with an invariant filter, enabling real-time visual-inertial odometry.

The position, orientation, and velocity of a robot system evolve on a matrix Lie group and possess symmetries (or invariance) in the sense that certain transformations leave the system state unchanged. Barrau et al. \cite{barrau2016invariant} introduced an invariant EKF in which the estimation errors remain invariant under the action of a matrix Lie group. Hartley et al. \cite{hartley2020contact} showed improved convergence and robustness of the invariant EKF in contact-aided inertial navigation, even when including bias terms in the filter state. Lin et al. \cite{lin2023proprioceptive} extend the latter work by developing an invariant state estimation approach using only onboard proprioceptive sensors. However, the inclusion of bias terms within the filter state breaks the Lie group symmetry, causing the linearized error dynamics to depend on the state estimates rather than remaining state-independent. Fornasier et al. \cite{fornasier2023msceqf} introduced an equivariant filter for VIO that integrates IMU bias and camera intrinsic-extrinsic parameters into a symmetry group structure. The approach achieves state-of-the-art accuracy and consistent estimation without the need for additional consistency enforcement techniques, e.g., observability constraint \cite{huang2010observability}. The equivariant filter extends the invariant filter by operating on homogeneous spaces, reducing to the invariant case with a specific choice of symmetry \cite{fornasier2023equivariant}. In this context, invariance corresponds to symmetries that leave the system state unchanged, whereas equivariance involves symmetries that change it in a structured manner \cite{fornasier2024equivariant}.

Our contribution is a sequence-to-sequence neural network that predicts IMU biases directly from past inertial measurements, which enables three key capabilities. First, estimating the bias outside the filter state enables an invariant Kalman filter, whose state covariance evolution is independent of the state estimates. Second, the proposed method achieves real-time visual inertial odometry. Third, we demonstrate the performance of the bias prediction network in visually degraded scenarios where the system relies solely on IMU measurements for motion estimation. Our evaluation demonstrates that this approach yields bias estimates that are physically consistent and stable (unlike the fluctuating estimates obtained from a standard EKF), and achieves reliable state estimation, even with poor or no visual features for extended periods of time.

\section{PROBLEM STATEMENT}

Consider a robot equipped with an IMU and a camera. The IMU provides noisy measurements of angular velocity $\omega(t) \in \bbR^3$ and linear acceleration $a(t) \in \bbR^3$. The camera provides the pixel coordinates $z(t) \in \mathbb{R}^2$ of keypoints tracked across consecutive images. IMU and camera measurements are assumed to be generated synchronously at the same discrete time steps $t_k$. 

Our goal is to estimate the robot's state at time $t$:
\begin{equation}
    \label{eq:robot_state}
    X(t) = \begin{bmatrix}
        R(t) & v(t) & p(t) \\ 
        0 & 1 & 0 \\ 
        0 & 0 & 1
    \end{bmatrix} \in SE_2(3), 
\end{equation}
where $R(t) \in SO(3)$, $v(t) \in \mathbb{R}^3$, and $p(t) \in \mathbb{R}^3$ denote the orientation, linear velocity, and position of the inertial frame relative to the global frame, respectively, and $SE_2(3)$ denotes the extended special Euclidean group \cite{brossard2021associating}.

The gyroscope and accelerometer measurements are corrupted by additive white noise $n^g(t), n^a(t) \in \bbR^3$ and time-varying bias $b^g(t),b^a(t)\in\bbR^3$, respectively:  
\begin{equation}
\begin{aligned}
    \bar{\omega}(t) &= \omega(t) + b^{g}(t) + n^{g}(t), \\ 
    \bar{a}(t) &= a(t) - R^{\top}(t) \; g + b^{a}(t) + n^{a}(t),
\end{aligned}
\end{equation}
where $\bar{\omega}(t) \in \bbR^3$ is the measurement of angular velocity in body-frame coordinates, $\bar{a}(t) \in \bbR^3$ is the measurement of linear acceleration in body-frame coordinates, and $g \in \bbR^3$ is the gravity vector in world-frame coordinates. Let $u(t) = (\omega(t), a(t))$ the denote noiseless measurements, $\bar{u}(t) = (\bar{\omega}(t), \bar{a}(t))$ denote the noisy measurements, and $b(t) = (b^{g}(t), b^{a}(t))$ denote the IMU bias. 

The evolution of state $X(t)$ with input $u(t)$ is governed by a continuous-time motion model:
\begin{equation}
    \label{eq:X_dot}
    \dot{X}(t) = f(X(t), u(t)) =\! 
    \begin{bmatrix}
        R(t) (\omega(t))_{\times} & R(t) a(t) + g & v(t) \\
        0 & 0 & 0 \\ 
        0 & 0 & 0
    \end{bmatrix}
\end{equation}
where the operator $(\cdot)_{\times}: \mathbb{R}^3 \rightarrow \mathfrak{so}(3)$ maps a vector in $\mathbb{R}^3$ to a $3 \times 3$ skew-symmetric matrix.

The IMU bias is typically modeled using a Brownian motion model (i.e., random walk) \cite{buchanan2022deep}: 
\begin{equation}
    \dot{b}(t) = \eta(t), \quad \eta(t) = 
    \begin{bmatrix}
        \eta^{g}(t)^{\top} & \eta^{a}(t)^{\top}
    \end{bmatrix}^{\top} \in \mathbb{R}^{6},
\end{equation}
where $\eta$ is the IMU bias noise. While this assumption provides a simple linear approximation of bias evolution, it might fail to capture complex behaviors.  Instead, we consider learning a sequence-to-sequence parametrized model $d_{\theta}$ that maps a sequence of IMU measurements to their corresponding sequence of biases, offering a more expressive model than a random walk.  
To achieve this, given a set of raw measurements $\bar{u}_{0:N}^{(i)}$, we predict the corresponding IMU biases $\hat{b}_{0:N}^{(i)}$ using $d_\theta$ and roll out the IMU kinematics $f$ in Eq. \eqref{eq:X_dot} with initial state $X_{0}^{(i)}$ and corrected measurements $\bar{u}_{k}^{(i)} - \hat{b}_{k}^{(i)}$, for $k = 0, \ldots, N$. 
We assume both the IMU measurements $u(t)$ and bias $b(t)$ remain constant during the time interval $[t_k, t_{k+1})$.

\begin{problem}\label{problem} 
Given dataset $\calD = \{t_{0:N}^{(i)}, X_{0:N}^{(i)}, \bar{u}_{0:N}^{(i)} \}_{i=1}^{D}$, learn an IMU bias prediction model $d_\theta$ by determining the parameters $\theta$ that minimize the following:
\begin{align}
\label{eq:problem_statement}
    \min_{\theta} \quad & \sum_{i=1}^{D} \sum_{k=1}^{N}  c( \hat{X}_{k}^{(i)}, X_{k}^{(i)}) \\
\textrm{s.t.} \quad & \hat{X}_{k+1}^{(i)} = \text{ODESolver}(f, \hat{X}_{k}^{(i)}, \bar{u}_{k}^{(i)} - \hat{b}_{k}^{(i)}, t_{k+1}^{(i)} - t_{k}^{(i)})  \nonumber \\
  \quad & \hat{b}_{0:N}^{(i)}  = d_{\theta}(\bar{u}_{0:N}^{(i)}), \; \text{for} \;  k = 0, \ldots, N \; \text{and} \; i = 1, \ldots, D,  \nonumber
\end{align}
for a given initial state $\hat{X}_{0}^{(i)} = X_{0}^{(i)}$. The cost function $c$ may be chosen as a suitable distance metric on the $SE_2(3)$ manifold. Instead of using an ODE solver (e.g., Runge-Kutta \cite{dormand1980family}), we compute the integration of Eq. \eqref{eq:X_dot} on the $SE_2(3)$ group in closed-form (shown in Sec.~\ref{sec:msckf}, Eq.~\eqref{eq:discrete_kinematic_equations}).
\end{problem}

\section{PRELIMINARIES}
\label{sec:preliminaries}

This section introduces background material that will be used throughout the paper.

\subsection{Lie Group Operators}

Let $X$ denote an element of the extended special Euclidean Lie group $SE_2(3)$, with structure defined in Eq. \eqref{eq:robot_state}. The corresponding Lie algebra $\frak{se}_2(3)$ consists of $9 \times 9$ matrices:
\begin{align}
    \xi^{\wedge} = 
    \begin{bmatrix}
        (\xi^{R})_{\times} & \xi^{v} & \xi^{p} \\ 
        0 & 0 & 0 \\ 
        0 & 0 & 0
    \end{bmatrix}, \; \xi =  
    \begin{bmatrix}
        \xi^{R} \\ \xi^{v} \\ \xi^{p}
    \end{bmatrix},
    \; \xi^{R}, \xi^{v}, \xi^{p} \in \mathbb{R}^3.
\end{align}
The vector $\xi \in \mathbb{R}^9$ parametrizes the Lie algebra via the hat operator $(\cdot)^{\wedge}: \mathbb{R}^9 \rightarrow \mathfrak{se}_2(3)$, while the vee operator $(\cdot)^{\vee}: \mathfrak{se}_2(3) \rightarrow \mathbb{R}^9$ is its inverse. A group element $X \in SE_2(3)$ is related to an algebra element $\xi^{\wedge} \in \mathfrak{se}_2(3)$ through the exponential $\exp(\cdot): \mathfrak{se}_2(3) \rightarrow SE_2(3)$ and logarithm $\log(\cdot): SE_2(3) \rightarrow \mathfrak{se}_2(3)$ maps:
\begin{align}
    \label{eq:lie_operators}
    X = \exp(\xi^{\wedge}), \quad  \xi^{\wedge} = \log(X), 
\end{align}
where $\exp(\xi^{\wedge})$ admits a closed-form expression: 
\begin{equation*}
\begin{aligned}
    \exp{(\xi^{\wedge}}) &= 
    \begin{bmatrix}
        \Gamma_0{(\xi^{R})} & \Gamma_{1}(\xi^{R}) \xi^{v} & \Gamma_{1}(\xi^{R}) \xi^{p} \\ 
        0 & 1 & 0 \\ 
        0 & 0 & 1
    \end{bmatrix}, 
\end{aligned} 
\end{equation*}
where $\Gamma_0(\cdot)$ and $\Gamma_1(\cdot)$ denote the $SO(3)$ exponential map and left Jacobian, respectively (see \cite{hartley2020contact}).

We consider right-invariant error associated with left perturbation. The group error state $\tilde{X}$ and retraction $\xi \oplus \hat{X}$ are defined as follows:
\begin{align}
    \label{eq:error_state}
    \tilde{X} = X \hat{X}^{-1}, 
    \quad
    \xi \oplus \hat{X} = \exp{(\xi^{\wedge})} \hat{X},
\end{align}
where the vector $\xi$ represents a perturbation in $\frak{se}_2(3)$. For $X \in SE_2(3)$, the adjoint map is defined as $\text{Ad}_{X}(\xi^{\wedge}) = X \xi^{\wedge} X^{-1}$ and its matrix representation can be written as: 
\begin{align}
    \label{eq:adjoint_map}
    \text{Ad}_{X} = 
     \begin{bmatrix}
        R & 0 & 0 \\
        (v)_{\times} R & R & 0 \\
        (p)_{\times} R & 0 & R \\
    \end{bmatrix}.
\end{align}
Please refer to \cite{barfoot2024state} for further details.

\subsection{Multi-state Constraint Kalman Filter (MSCKF)}
\label{sec:msckf}
The MSCKF \cite{mourikis2007multi} is a VIO method that marginalizes landmark positions instead of incorporating them in the filter state, thereby avoiding to build a map of 3D landmark positions. The MSCKF maintains a sliding window of past sensor poses to triangulate keypoints via least-squares optimization using geometric constraints from multiple images.

The filter state consists of the robot state $X_k \in SE_2(3)$, IMU bias $b_k$ at time $t_k$, and a window of $W$ historical states $X_{k-1} \ldots, X_{k-W}$. Given inertial measurement $\bar{u}_k$, the mean of the state $\hat{X}(t)$ and of the bias $\hat{b}(t)$ are propagated as: 
\begin{align}
    \label{eq:msckf_nonlinear_dynamics}
    \dot{\hat{X}}(t) = f(\hat{X}(t), \bar{u}_k - \hat{b}(t)), \quad \dot{\hat{b}}(t) = 0.
\end{align}
In the remainder of the paper, we omit the time dependence of the variables for readability. We denote estimated quantities with $\hat{(\cdot)}$ and error quantities with $\tilde{(\cdot)}$. The MSCKF uses decoupled error states $\exp(\xi^{R}_{\times}) = R \hat{R}^{\top}$,  $\xi^{v} = v - \hat{v}$, $\xi^{p} = p - \hat{p}$, and $\tilde{b} = b - \hat{b}$ to propagate the IMU covariance $P \in \mathbb{R}^{15 \times 15}$ through the linearized error-state dynamics: 
\begin{equation}\label{eq:error_dynamics}
    \begin{bmatrix}
        \dot{\xi} \\ \dot{\tilde{b}} 
    \end{bmatrix} = A
    \begin{bmatrix}
        \xi \\ \tilde{b}
    \end{bmatrix} + G
    \begin{bmatrix}
        n \\ \eta
    \end{bmatrix},  
\end{equation}
where $n = \begin{bmatrix} n^{g \top} & n^{a \top} & n^{v \top} \end{bmatrix} \in \mathbb{R}^{9}$ is the noise associated with angular velocity, linear acceleration, linear velocity. Here, $A$ and $G$ represent the Jacobians resulting from linearizing the error-state dynamics around the filter state estimate \cite{mourikis2007multi}. 
Thus, the IMU covariance evolution is governed by the continuous-time Riccati equation \cite{hartley2020contact}: 
\begin{align}
    \label{eq:riccati_equation}
    \dot{P} &= A P + P A^{\top} + Q, \quad Q = G \mathrm{Cov}(n) G^{\top}.
\end{align}

\paragraph*{Filter propagation}
To propagate the means of the state $\hat{X}_k$ and of the bias $\hat{b}_k$ between $t_k$ and $t_{k+1}$, with the assumption that the IMU measurement $\bar{u}_k$ remains constant over the time interval $\Delta t_k =  t_{k+1} - t_k$, \cite{hartley2020contact} provides closed-form integration of Eq. \eqref{eq:msckf_nonlinear_dynamics}: 
\begin{align}
    \label{eq:discrete_kinematic_equations}
    \hat{R}_{k+1} &= \hat{R}_k \Gamma_{0}{((\bar{\omega}_k - \hat{b}_{k}^{g}) \Delta t_k)}, \nonumber \\ 
    \hat{v}_{k+1} &= \hat{v}_k + g \Delta t_k + \hat{R}_k \Gamma_{1}{((\bar{\omega}_k - \hat{b}_{k}^{g}) \Delta t_k)} (\bar{a}_k - \hat{b}_{k}^{a}), \nonumber \\ 
     \hat{p}_{k+1} &= \hat{p}_k + \hat{v}_k \Delta t_k + \frac{1}{2} g \Delta t_k^2 \nonumber \\ 
     & + \hat{R}_k \Gamma_{2}{((\bar{\omega}_k - \hat{b}_{k}^{g}) \Delta t_k)} (\bar{a}_k - \hat{b}_{k}^{a}), 
\end{align}
where $\Gamma_2(\phi)$ is defined in \cite{hartley2020contact}. To obtain the covariance of $\hat{X}_k$ and $\hat{b}_k$, we approximate the integration of Eq. \eqref{eq:riccati_equation}: 
\begin{equation} \label{eq:discrete_covariance_propagation}
\begin{aligned}
    P_{k+1} &= \Phi_{k} P_{k} \Phi_{k}^{\top} + Q_{k}^{d}, \qquad
    \Phi_{k} = \exp(A_k \Delta t_k), \\ 
    Q_{k}^{d} &\approx  \Phi_{k} Q_k \Phi_{k}^{\top} \Delta t_k, \quad \quad Q_k = G_k \mathrm{Cov}(n) G_k^{\top}.
\end{aligned}
\end{equation} 
Since past states remain constant, their covariance entries are propagated using an identity Jacobian and zero process noise, as described in \cite{geneva2020openvins}. 


\paragraph*{Filter update} 
Consider a keypoint $z_{k,m} \in \mathbb{R}^{2}$, obtained from an image keypoint detection algorithm such as FAST \cite{rosten2008faster}, associated with landmark $\ell_m \in \mathbb{R}^3$ and state $X_k$. The variables are related by the measurement model \cite{shan2020orcvio}:
\begin{equation}\label{eq:camera_model}
z_{k,m} = h(X_k, \ell_m) + \rho_{k,m},  
\end{equation}
where $h$ is the image projection of landmark $\ell_m$ and $\rho_{k,m}$ is the keypoint detection noise. Define the keypoint error for each measurement as $e_{k,m} = z_{k,m} - h(\hat{X}_k, \hat{\ell}_m)$. After applying the left null-space projection step from \cite{mourikis2007multi}, let $\hat{e}$, $H$, and $V$ represent the stacked errors, measurement Jacobians, and noise covariances for all landmarks. We update the mean $\hat{\calX}_k = (\hat{X}_k, \hat{b}_k, \hat{X}_{k-1}, \ldots, \hat{X}_{k-W})$, and covariance $\calP_k$ as: 
\begin{equation}
\begin{aligned}
    \hat{\calX}_{k+1} &= (K \hat{e}) \oplus \hat{\calX}_{k+1},\\ 
    \calP_{k+1} &= (I - KH) \calP_{k+1} (I - KH)^{\top} + KVK^{\top},  \\ 
    K &= \calP_k H^{\top} \left( H \calP_k H^{\top} + V \right)^{-1}.
\end{aligned}
\end{equation}

\section{Learning IMU Bias for VIO}
\label{sec:learning_bias}
Typically, the IMU bias is not directly observable from a single IMU measurement. Instead, bias estimation requires integrating a sequence of IMU measurements and comparing against an external sensor for ground truth (e.g., motion capture system). This sequence dependence motivates us to predict IMU biases from a sequence of raw IMU measurements via a neural network model and using the ground-truth state for training. If ground-truth poses are unavailable, a camera sensor can be used to track keypoints across frames to estimate relative poses $X_{k+1}^{-1} X_k$, which can be used as ground-truth. Recent works propose using deep learning architectures, such as Convolutional Neural Network (CNN), ResNet, or transformer, to capture temporal dependencies and patterns in IMU measurements \cite{qiu2023airimu, buchanan2022deep, liu2020tlio, chen2018ionet, cioffi2023learned, herath2020ronin, zhang2022dido}. In this work, we learn a sequence-to-sequence neural network model $d_{\theta}$, mapping a sequence of IMU measurements $\bar{u}_{k-L}^{(i)}, \ldots, \bar{u}_{k}^{(i)}$ to a corresponding sequence of IMU bias estimates $(\hat{b}_{k-L}^{(i)}, \ldots, \hat{b}_{k}^{(i)})$:  
\begin{equation}
\label{eq:neural_network_model}
    d_{\theta}(\bar{u}_{k-L}^{(i)}, \ldots, \bar{u}_{k}^{(i)}) = (\hat{b}_{k-L}^{(i)}, \ldots, \hat{b}_{k}^{(i)}). 
\end{equation}
Due to the correlation between angular velocity and acceleration at slow-varying velocities, e.g., observed in \cite{brossard2020denoising}, we use a single model $d_\theta$ to infer the IMU bias instead of separate models for the gyroscope and accelerometer.

To optimize $\theta$, we partition the collected trajectories into non-overlapping $D$ segments, each consisting of $N$ samples of ground-truth state $X_k^{(i)}$ and raw IMU measurements $\bar{u}_k^{(i)}$. For each segment, we feed the raw IMU measurements into the neural network model $d_{\theta}$ to predict the corresponding bias estimates $\hat{b}_{0:N}^{(i)}$ as in Eq. \eqref{eq:neural_network_model}, which are expected to initially be inaccurate. These predicted biases are then used to correct the raw IMU measurements with $\bar{u}_k^{(i)} - \hat{b}_{0:N}^{(i)}$. Then, we roll out an estimated state trajectory $\hat{X}_{1:N}^{(i)}$ with the corrected IMU measurements $\bar{u}_k^{(i)} - \hat{b}_{0:N}^{(i)}$ and an initial state $X_0^{(i)}$ using Eq. \eqref{eq:discrete_kinematic_equations}. To update the neural network parameters $\theta$, we use a cost function $c( \hat{X}_{k}^{(i)}, X_{k}^{(i)})$ that measures the discrepancy between the predicted state $\hat{X}_k^{(i)}$ and the ground-truth $X_k^{(i)}$. \NEW{As states lie on $SE_2(3)$}, we compute the group error $\tilde{X}_k^{(i)} = X_k^{(i)} \hat{X}_k^{(i)^{-1}} \in SE_2(3)$, map it onto the Lie algebra $\xi_k^{(i) \wedge} = \log(\tilde{X}_k^{(i)}) \in \frak{se}_2(3)$, and calculate the norm of its \NEW{vector representation $\xi_k^{(i)} = \log(\tilde{X}_k^{(i)})^{\vee} \in \mathbb{R}^9$} as follows:
\begin{equation} \label{eq:loss_function}
     c( \hat{X}_{k}^{(i)}, X_{k}^{(i)}) = \left\| \log{(X_k^{(i)} \hat{X}_k^{(i)^{-1}})^{\vee}} \right\|_{h}.
\end{equation} 
We use the Huber loss $\left\| \cdot \right\|_{h}$ in the cost function definition above to prioritize early trajectory estimates, which are less corrupted by the noise inherent in the IMU measurements which accumulates with long-horizon integration of Eq. \eqref{eq:discrete_kinematic_equations}. Alternatively, the network can be trained with a loss on relative poses $X_{k+1}^{-1} X_k$, which delivers similar results. The 9-dimensional state error vector evaluated by the Huber loss is weighted by $10^3$ for orientation, $10^2$ for position, and $10^1$ for velocity to give the three components equal influence despite their different units. We use the Adam optimizer \cite{kingma2017adam} with learning rate $1 \times 10^{-3}$ to iteratively optimize the parameters $\theta$. Our approach provides high-frequency bias estimates compared to updating the bias only at the lower-frequency update step, typically at the camera frame rate. In addition, our model decouples the bias prediction from visual information, which can be unreliable for extended periods.

\begin{figure}[t]
    \centering
    \includegraphics[width=1.0\linewidth]{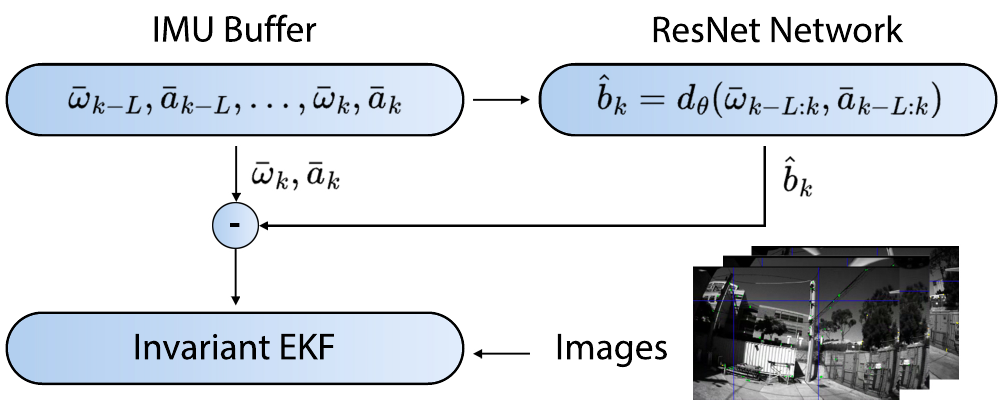}
    \caption{\NEW{Block diagram of our system.}}
    \label{fig:block_diagram}
    \vspace{-5mm}
\end{figure}

We use the invariant EKF \cite{barrau2016invariant} to track the filter state mean $\hat{X}_k, \ldots, \hat{X}_{k-W}$ on the matrix Lie group $SE_2(3)$, while the covariance $P_k$ is propagated in the corresponding Lie algebra. 
Relying on learned bias predictions from $d_\theta$ allows us to formulate the IMU dynamics without introducing the bias in the filter state: 
\begin{equation}
    \dot{X} = f(X, \bar{u}_k - b) - X n^{\wedge},
\end{equation}
where $n$ is the propagation noise as in Eq.~\eqref{eq:error_dynamics}. From \cite{barrau2016invariant}, the deterministic system $f$ satisfies the group-affine property. Thus, using the group error defined in Eq. \eqref{eq:error_state}, the linearized IMU error-state dynamics:
\begin{align}
\label{eq:linearized_error_dynamics_in_lie_algebra}
    \dot{\xi}
     &= A\xi +  \text{Ad}_{X} n, \quad  
    A = 
    \begin{bmatrix}
        0 & 0 & 0 \\
        (g)_{\times} & 0 & 0 \\
        0 & I & 0
    \end{bmatrix}, 
\end{align}
can be propagated as in Eq. \eqref{eq:riccati_equation} with $G = \text{Ad}_{X}$. Note that for the deterministic system $\dot{\xi} = A \xi$, since the Jacobian $A$ is state-independent, the error propagation is independent of the state estimate. When noise is brought to the system, the state covariance mapping remains state-independent, whereas the noise mapping depends on the state estimate as in Eq. \eqref{eq:discrete_covariance_propagation}, which is an advantage over the standard EKF. To summarize, \NEW{as shown in Fig.~\ref{fig:block_diagram},} for a given state mean $\hat{X}_k$ and covariance $P_{k}$ with raw IMU measurements $\bar{u}_{k-L}, \ldots, \bar{u}_{k}$, we have:
\begin{align}
    (\hat{b}_{k-L}, \ldots, \hat{b}_{k}) &= d_{\theta}(\bar{u}_{k-L}, \ldots, \bar{u}_{k}), \notag \\ 
    \Phi_{k} &= \exp(A \Delta t_k), \\ 
    P_{k+1} &= \Phi_{k} P_k \Phi_{k}^{\top} + \Phi_{k} \text{Ad}_{X_k} \mathrm{Cov}(n) \text{Ad}_{X_k}^{\top} \Phi_{k}^{\top} \Delta t_k \notag, \\
    \text{and}& \quad  \text{ Eq. \eqref{eq:discrete_kinematic_equations}}. \notag
\end{align}

\begin{figure*}[!t]
    \subcaptionbox{Position \label{fig:position}}{\includegraphics[width=0.33\linewidth,trim={5pt 15pt 5pt 5pt},clip]{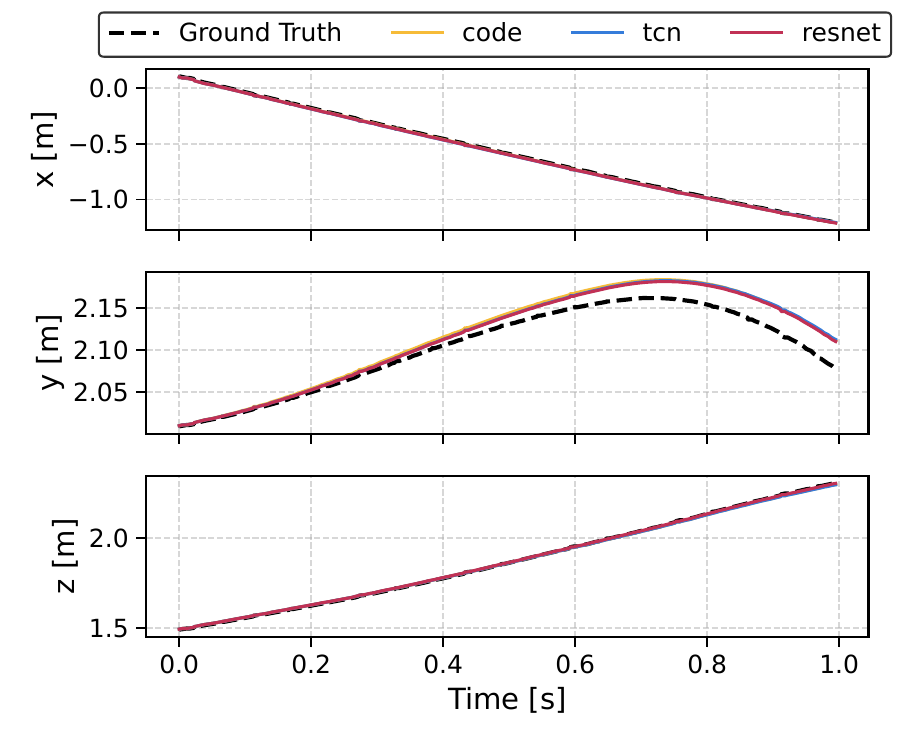}}
    \subcaptionbox{Orientation \label{fig:orientation}}{\includegraphics[width=0.33\linewidth,trim={5pt 15pt 5pt 5pt},clip]{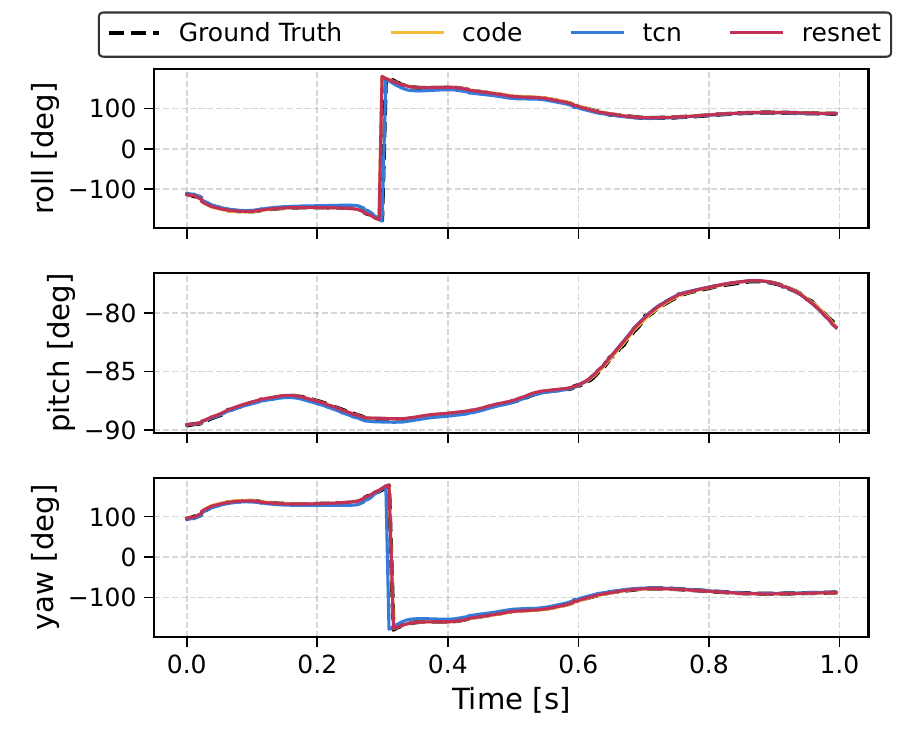}} 
    \subcaptionbox{Velocity \label{fig:velocity}}{\includegraphics[width=0.33\linewidth,trim={5pt 15pt 5pt 5pt},clip]{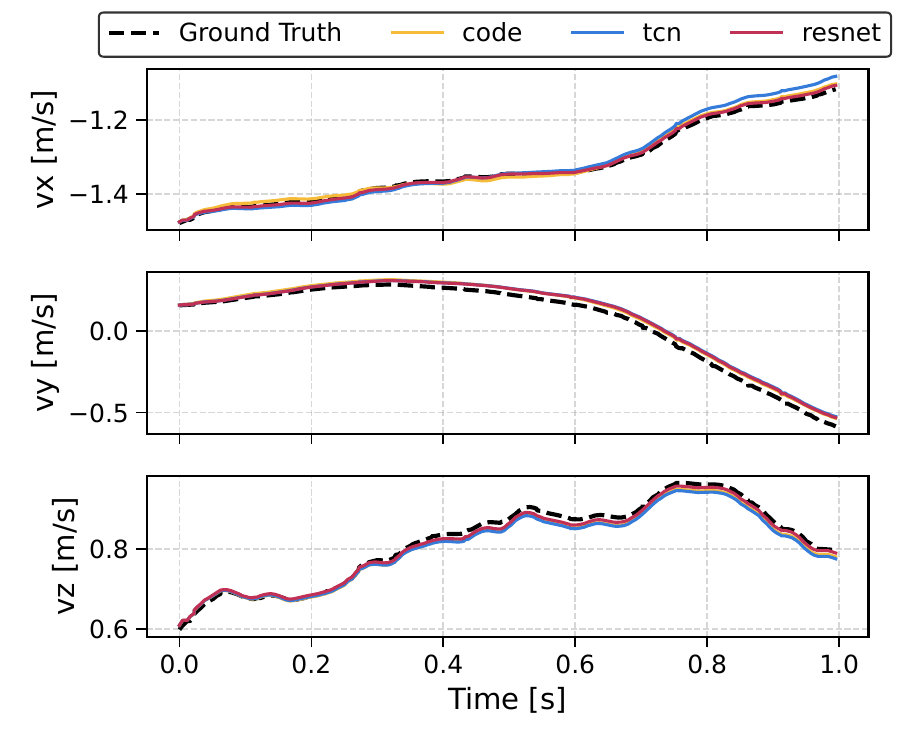}}  
    \caption{Predicted position, orientation, and velocity over a 1-second window from initial state $X_0$ on the \emph{Aerodrome} dataset, comparing CodeNet, TCN, and ResNet predictions against ground-truth.}
    \label{fig:simulation_results}
\end{figure*}

\begin{figure*}[!t]
    \centering
    \includegraphics[width=1.0\linewidth]{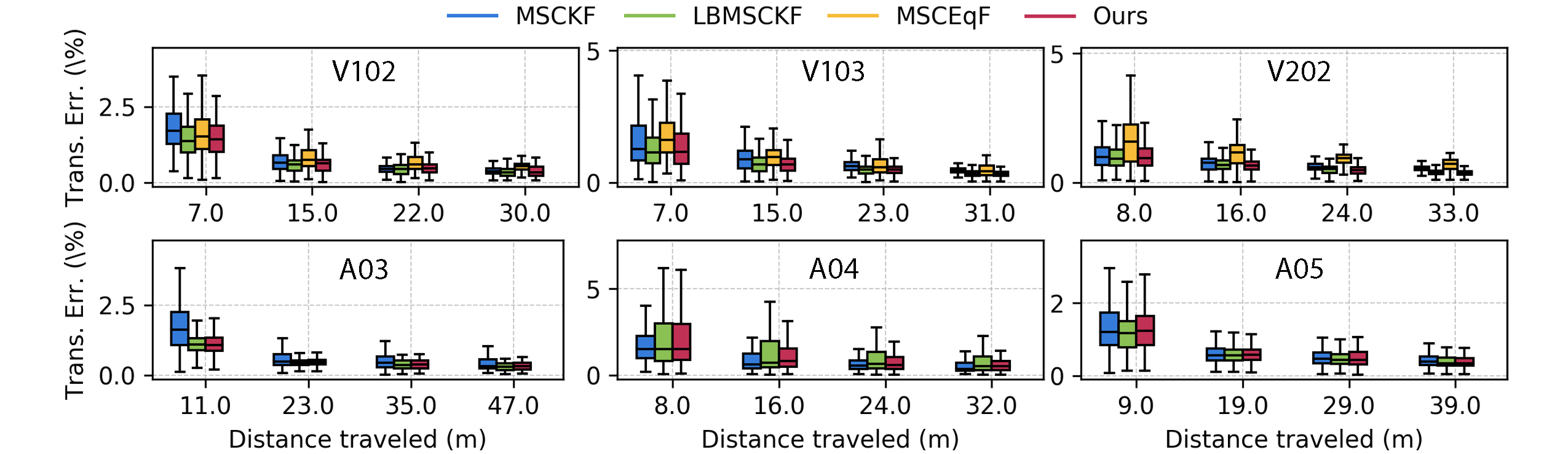}
    \caption{\NEW{RE on \emph{EuRoC} and \emph{Aerodrome} over 2.5\%, 5\%, 7.5\%, and 10\% traveled distances of the total trajectory, averaged over five runs.}}
    \label{fig:euroc_combined}
\end{figure*}

\section{Evaluation}
We present our choice of neural network architecture for IMU bias prediction in Sec. \ref{sec:model_arch}. Then, we evaluate our method against state-of-the-art VIO baselines using both publicly available dataset and \emph{Aerodrome} dataset with challenging motions in Sec. \ref{sec:euroc}. We demonstrate the robustness of our method under challenging conditions where visual features are temporarily lost, requiring the filter to rely solely on IMU measurements, in Sec. \ref{sec:extreme_scenarios}. Finally, in Sec. \ref{sec:inertial_odometry}, we evaluate our method as an inertial-only odometry approach and compare it against two popular inertial-odometry baselines. 

\paragraph*{Datasets} We evaluate on the public \emph{EuRoC} dataset \cite{burri2016euroc} and our \emph{Aerodrome} dataset. The \emph{EuRoC} dataset provides $200$ Hz IMU, $20$ Hz camera, and $100$ Hz ground truth from a quadrotor operating at maximum speeds of $2.3$ m/s. \NEW{Per \cite{brossard2020denoising}, MH and VR1 were captured on consecutive days, while VR2 was acquired later}. Thus, we train on (MH01, MH03, MH04), validate on (MH02, MH05), and test on (V102, V103, V202). The \emph{Aerodrome} dataset consists of five trajectories with $200$ Hz IMU data, $25$ Hz camera, and $100$ Hz ground truth from a quadrotor operating at maximum speeds of $5.4$ m/s. We train on A01, validate on A02, and test on (A03, A04, A05).

\paragraph*{Metrics} To assess IMU bias prediction accuracy in Sec. \ref{sec:model_arch}, we compute the cost defined in Eq. \eqref{eq:problem_statement} over the test data along with the average error norms of the orientation $\Vert \xi^{R} \Vert = \Vert \log(\hat{R}R^\top)^{\vee} \Vert$, velocity $\Vert \xi^{v} \Vert = \Vert \hat{v} - v \Vert$, and position $\Vert\xi^{p}\Vert = \Vert \hat{p} - p \Vert$. For quantitative trajectory evaluation in Sec. \ref{sec:euroc}, \ref{sec:extreme_scenarios}, and \ref{sec:inertial_odometry}, we report the Absolute Trajectory Error (ATE) in translation and rotation and the Relative Error (RE) in translation. \NEW{These metrics are defined in \cite{zhang2018tutorial}}.

\paragraph*{Baselines} We evaluate our approach against three VIO methods: MSCKF \cite{geneva2020openvins}, a monocular multi-state constraint Kalman filter with IMU bias estimated in the filter state, MSCEqF \cite{fornasier2023msceqf}, an equivariant formulation of the monocular MSCKF with IMU bias estimated in the filter state, and LBMSCKF, an MSCKF with learned bias. In addition, we evaluate our approach as an inertial-only odometry against two learning-based inertial odometry methods: IMO \cite{cioffi2023learned}, which combines a TCN to estimate relative positions from IMU and thrust inputs with an EKF, and TLIO \cite{liu2020tlio}, which uses a ResNet to estimate relative position displacements in a local gravity-aligned frame and its uncertainty from IMU measurements. In both methods, the IMU bias is estimated within the filter state.

\subsection{Model Architecture Choice and Implementation Details}
\label{sec:model_arch}

We investigate the choice of neural network architecture for IMU bias prediction that obtains the best state prediction on the \emph{EuRoC} and \emph{Aerodrome} datasets. Motivated by the sequential dependence in Sec. \ref{sec:learning_bias}, we compare three commonly used sequential architectures: ResNet \cite{liu2020tlio}, TCN \cite{cioffi2023learned}, and CodeNet \cite{qiu2023airimu} as candidate models for $d_\theta$.

\paragraph*{Implementation Details} The neural network architectures compared in this evaluation have a comparable number of trainable parameters: $300K$ for ResNet, $500K$ for TCN, and $400K$ for CodeNet. Given a sequence of IMU measurements sampled at $200$ Hz over a one-second window ($L=200$), along with an initial state $X_0$, each network estimates biases $\hat{b}_{k-L}, \ldots, \hat{b}_{k}$, as in Eq. \eqref{eq:neural_network_model}, and corrects for the raw measurements before integration as in Eq. \eqref{eq:discrete_kinematic_equations}. The ResNet follows its original design, predicting a single bias estimate assumed constant $\hat{b}_k = \hat{b}_i$ for all $i = k-L,\ldots,k$ throughout the prediction window. The implementation of TCN in \cite{cioffi2023learned} outputs a 3-dimensional vector. We modified the TCN architecture to predict a sequence of 6-dimensional output, representing both gyroscope and accelerometer biases. In our implementation, the filter state maintains up to $W=11$ past states, enabling real-time operation with filter update steps at $20$ Hz, while neural network inference runs at $200$ Hz on every incoming IMU measurement using an overlapping one-second sliding window. Table~\ref{tab:noise_parameters} presents the IMU noise parameters used for the filters in all experiments. The computation times, recorded onboard the quadrotor with an Intel i7 NUC, are listed in Table~\ref{tab:computation_time}, showing that bias inference adds negligible overhead to the total processing time.

\begin{table}[!t]
	\centering
	\caption{IMU Noise Parameters}
	\begin{tabular}{ccccc}
		\hline
		Parameter & Symbol & EuRoC & Aerodrome & Unit\\
		\hline
            Gyro. Noise Density & $\sigma_{g}$ & $1e^{-2}$ & $1e^{-2}$  & $\frac{\text{rad}}{\text{s}}\frac{1}{\sqrt{Hz}}$ \\[.5ex]
            Gyro. Random walk & $\sigma_{bg}$ & $8e^{-4}$ & $6e^{-4}$ & $\frac{\text{rad}}{\text{s}^2}\frac{1}{\sqrt{Hz}}$ \\[.5ex]
            Accel. Noise Density & $\sigma_{a}$ & $3e^{-2}$ & $1e^{-1}$ & $\frac{\text{m}}{\text{s}^2}\frac{1}{\sqrt{Hz}}$ \\[.5ex]
            Accel. Random Walk & $\sigma_{ba}$ & $2e^{-4}$ & $7e^{-3}$ & $\frac{\text{m}}{\text{s}^3}\frac{1}{\sqrt{Hz}}$ \\[.5ex]
		\hline 
		\label{tab:noise_parameters}
	\end{tabular}
    \vspace{-5mm}
\end{table}

\NEW{
\begin{table}[!t]
    \caption{IMU bias learning with different network architectures.}
    \label{table:compare_net_arch}
    \centering
  \begin{adjustbox}{width=\columnwidth,center}
	\begin{tabular}{ccccc} 
		\textbf{Metrics} & \textbf{Dataset} & \textbf{TCN} & \textbf{CodeNet} & \textbf{ResNet} \\
		\hline
		\hline
  		Test loss in Eq. \eqref{eq:loss_function} & EuRoC & $0.022$  & $0.023$ & ${\bf 0.022}$  \\
            $\Vert \log(\hat{R}R^\top)^{\vee} \Vert$ (avg.) & EuRoC & $1.76 \times 10^{-6}$ & ${\bf 1.70 \times 10^{-6}}$ & $2.3 \times 10^{-6}$ \\
            $\Vert \hat{v} - v \Vert$ (avg.) & EuRoC & $7.46 \times 10^{-4}$ & $8.01 \times 10^{-4}$ & ${\bf 7.26 \times 10^{-4}}$ \\
            $\Vert \hat{p} - p \Vert$ (avg.) & EuRoC & $1.27 \times 10^{-4}$ & $1.35 \times 10^{-4}$ & ${\bf 1.24 \times 10^{-4}}$
            \\\hline
		Test loss in Eq. \eqref{eq:loss_function} & Aerodrome & $0.013$ & ${\bf 0.005}$ & ${\bf 0.005}$ \\
            $\Vert \log(\hat{R}R^\top)^{\vee} \Vert$ (avg.) & Aerodrome & $6.88 \times 10^{-3}$ & ${\bf 1.69 \times 10^{-3}}$ & $2.27 \times 10^{-3}$ \\
            $\Vert \hat{v} - v \Vert$ (avg.) & Aerodrome & $2.37 \times 10^{-2}$ & $2.66 \times 10^{-2}$ & ${\bf 2.32 \times 10^{-2}}$ \\
            $\Vert \hat{p} - p \Vert$ (avg.) & Aerodrome & $1.12 \times 10^{-2}$ & ${\bf 1.03 \times 10^{-3}}$ & ${\bf 1.03 \times 10^{-3}}$
            \\\hline
	\end{tabular}
  \end{adjustbox}
\end{table}
}

\paragraph*{Training Results} Table~\ref{table:compare_net_arch} summarizes the state prediction accuracy achieved with bias correction using different network architectures, measured by the metrics defined earlier. ResNet shows average improvements in test, velocity, and position losses, while TCN achieves the best rotation estimation performance on average. We observe a comparable performance across the architectures, with ResNet showing a slight edge. A qualitative evaluation is provided in Fig. \ref{fig:simulation_results}, illustrating the predicted position, orientation, and linear velocity over a one-second interval starting from a known initial state $X_0$. Visually, these predictions align with the quantitative results reported in Table~\ref{table:compare_net_arch}. Therefore, we use the ResNet architecture to evaluate our IMU bias prediction for VIO in the remainder of the paper. In addition, we provide an ablation study on the window length $L$, reported in Table \ref{tab:window_length}. The error decreases with longer windows, so we set $L=200$ in our model for VIO evaluation.

\begin{table}[!t]
    \caption{Ablation on window size $L$ for \emph{Aerodrome} dataset.}
    \label{tab:window_length}
    \centering
  \begin{adjustbox}{width=\columnwidth,center}
	\begin{tabular}{cccccc} 
		\textbf{Metrics} & \textbf{Architecture} & $L=50$ & $L=100$ & $L=150$ & $L=200$ \\
		\hline
		\hline
            $\Vert \log(\hat{R}R^\top)^{\vee} \Vert$ (avg.) & & $1.1 \times 10^{-2}$ & $8.6 \times 10^{-3}$ & $7.8 \times 10^{-3}$ & $\NEW{\mathbf{6.9 \times 10^{-3}}}$ \\
            $\Vert \hat{v} - v \Vert$ (avg.) & TCN & $6.9 \times 10^{-2}$ & $4.7 \times 10^{-2}$ & $4.6 \times 10^{-2}$ & $\mathbf{2.4 \times 10^{-2}}$ \\
            $\Vert \hat{p} - p \Vert$ (avg.) &  & $7.5 \times 10^{-2}$ & $4.3 \times 10^{-2}$ & $3.9 \times 10^{-2}$ & $\mathbf{1.1 \times 10^{-2}}$
            \\\hline
            $\Vert \log(\hat{R}R^\top)^{\vee} \Vert$ (avg.) & & $8.9 \times 10^{-3}$ & $3.2 \times 10^{-3}$ & $1.8 \times 10^{-3}$ & $\mathbf{1.7 \times 10^{-3}}$ \\
            $\Vert \hat{v} - v \Vert$ (avg.) & CodeNet & $7.0 \times 10^{-2}$ & $4.8 \times 10^{-2}$ & $4.6 \times 10^{-2}$ & $\mathbf{2.7\times 10^{-2}}$ \\
            $\Vert \hat{p} - p \Vert$ (avg.) &  & $7.5 \times 10^{-2}$ & $4.4 \times 10^{-2}$ & $3.8 \times 10^{-2}$ & $\mathbf{1.0 \times 10^{-3}}$
            \\\hline
	    $\Vert \log(\hat{R}R^\top)^{\vee} \Vert$ (avg.) & & $1.1 \times 10^{-2}$ & $4.3 \times 10^{-3}$ & $2.3 \times 10^{-3}$ & $\mathbf{2.2 \times 10^{-3}}$ \\
            $\Vert \hat{v} - v \Vert$ (avg.) & ResNet & $7.0 \times 10^{-2}$ & $4.7 \times 10^{-2}$ & $4.5 \times 10^{-2}$ & $\mathbf{2.3 \times 10^{-2}}$ \\
            $\Vert \hat{p} - p \Vert$ (avg.) &  & $7.5 \times 10^{-2}$ & $4.3 \times 10^{-2}$ & $3.9 \times 10^{-2}$ & $\mathbf{1.0 \times 10^{-3}}$
            \\\hline
	\end{tabular}
  \end{adjustbox}
  \vspace{-5mm}
\end{table}

\subsection{Comparison to MSCKF and MSCEqF}
\label{sec:euroc}

We use the inertial noise parameters \NEW{in Table} \ref{tab:noise_parameters} and disable the SLAM features of the MSCKF baseline \cite{geneva2020openvins} for a fair comparison. We omitted MSCEqF from the \emph{Aerodrome} evaluation due to issues in the open-source code with the keypoint detector failing to extract image features. Table~\ref{tab:euroc_evaluations} and Fig. \ref{fig:euroc_combined} present results on the \emph{EuRoC} and \emph{Aerodrome} datasets using the ATE and RE metrics. MSCKF and LBMSCKF achieve nearly the same accuracy on \emph{EuRoC}, while on \emph{Aerodrome}, which includes maneuvers up to $5.4$ m/s, LBMSCKF outperforms MSCKF across all sequences. Our method achieves the best or second-best ATE results on all sequences, except on V102, thereby outperforming the baselines on average. Similarly, Fig. \ref{fig:euroc_combined} shows lower mean RE for our method across all sequences. In addition, we achieve lower variance errors compared to the baselines, suggesting better reliability. Overall, our method exhibits a slight edge in performance in normal scenarios. Both the learned bias prediction and the invariant filter formulation contribute complementary improvements in the performance. In the next subsection, we examine the robustness of our approach against the MSCKF under challenging conditions, where visual features are temporarily lost.

\begin{table}[t]
    \caption{\NEW{Evaluation on \emph{EuRoC} and \emph{Aerodrome}: best in bold, second-best underlined, averaged over five runs per sequence.}}
    \label{tab:euroc_evaluations}
    \centering
  \begin{adjustbox}{width=\columnwidth,center}
    \begin{tabular}{cccccc}
        \textbf{Metrics} & \textbf{Sequence} & \textbf{MSCKF \cite{geneva2020openvins}} & \textbf{MSCEqF \cite{fornasier2023msceqf}} & \textbf{LBMSCKF} & \textbf{Ours} \\
        \hline
        \hline
        ATE trans. [m] & V102 & ${\bf 0.111}$ & $0.140$ & $\underline{0.122}$ & $0.129$ \\
        ATE rot. [deg] & V102 & $3.931$ & ${\bf 1.470}$ & $2.877$ & $\underline{2.193}$ \\
        \hline
        ATE trans. [m] & V103 & $0.158$ & $0.164$ & $\underline{0.157}$ & ${\bf 0.140}$ \\
        ATE rot. [deg] & V103 & ${\bf 0.858}$ & $3.598$ & $1.834$ & $\underline{1.584}$ \\
        \hline
        ATE trans. [m] & V202 & $\underline{0.148}$ & $0.182$ & $0.153$ & ${\bf 0.125}$ \\
        ATE rot. [deg] & V202 & $2.372$ & ${\bf 1.707}$ & $3.973$ & $\underline{2.198}$ \\
        \hline
        ATE trans. [m] & A03 & $0.220$ & - & ${\bf 0.142}$ & ${\bf 0.142}$ \\
        ATE rot. [deg] & A03 & $1.524$ & - & ${\bf 0.524}$ & $\underline{0.641}$ \\
        \hline
        ATE trans. [m] & A04 & $1.355$ & - & $\underline{0.742}$ & ${\bf 0.386}$ \\
        ATE rot. [deg] & A04 & $4.841$ & - & ${\bf 3.508}$ & $\underline{3.860}$ \\
        \hline
        ATE trans. [m] & A05 & $0.201$ & - & $\underline{0.191}$ & ${\bf 0.190}$ \\
        ATE rot. [deg] & A05 & $1.382$ & - & ${\bf 0.773}$ & $\underline{0.821}$ \\
        \hline
    \end{tabular}
  \end{adjustbox}
  \vspace{-5mm}
\end{table}

\subsection{Comparison to MSCKF in Extreme Scenarios}
\label{sec:extreme_scenarios}
\begin{figure}[t]
    \centering
    \includegraphics[width=1.0\linewidth]{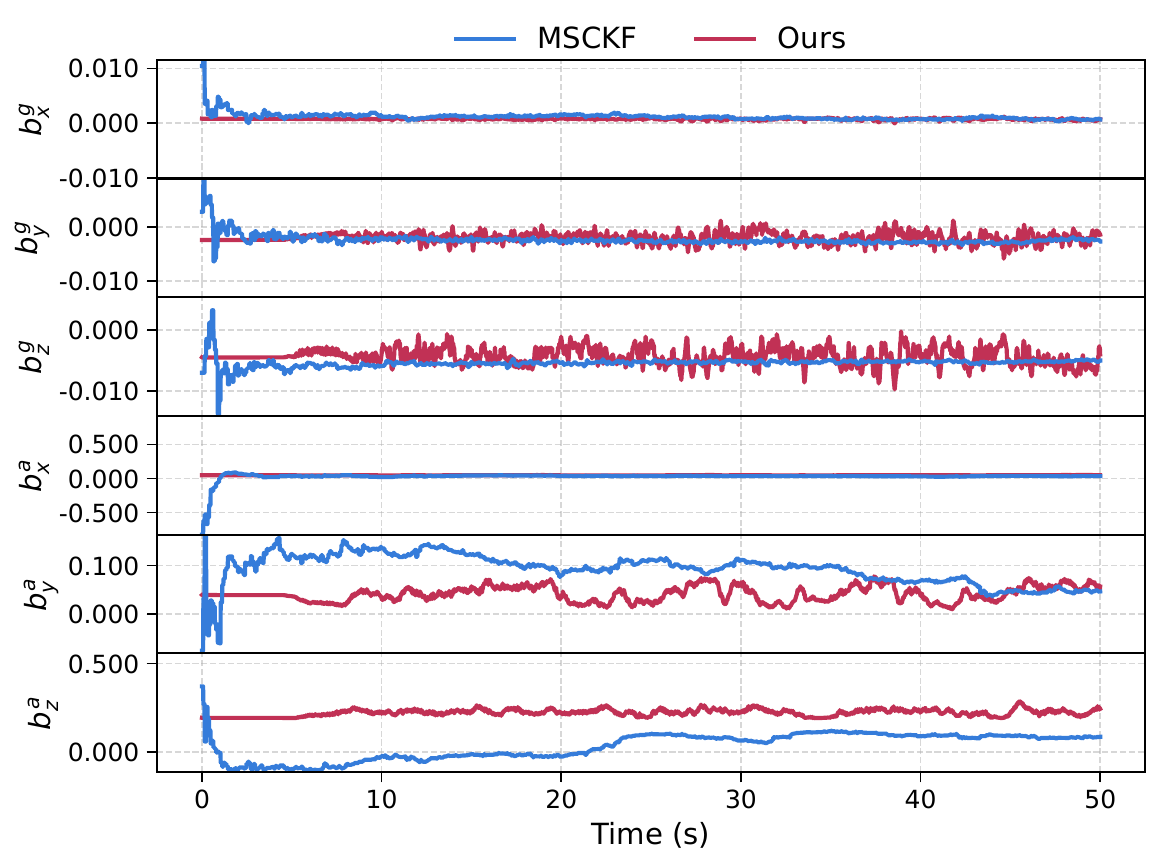}
    \caption{Comparison of IMU bias estimates over time on \emph{Aerodrome} A03 sequence between MSCKF (blue) and our network (red).}
    \label{fig:bias_evolution}
    \vspace{-5mm}
\end{figure}

\begin{figure}[t]
    \centering
    \includegraphics[width=1.0\linewidth, trim={0pt 8pt 0pt 0pt},clip]{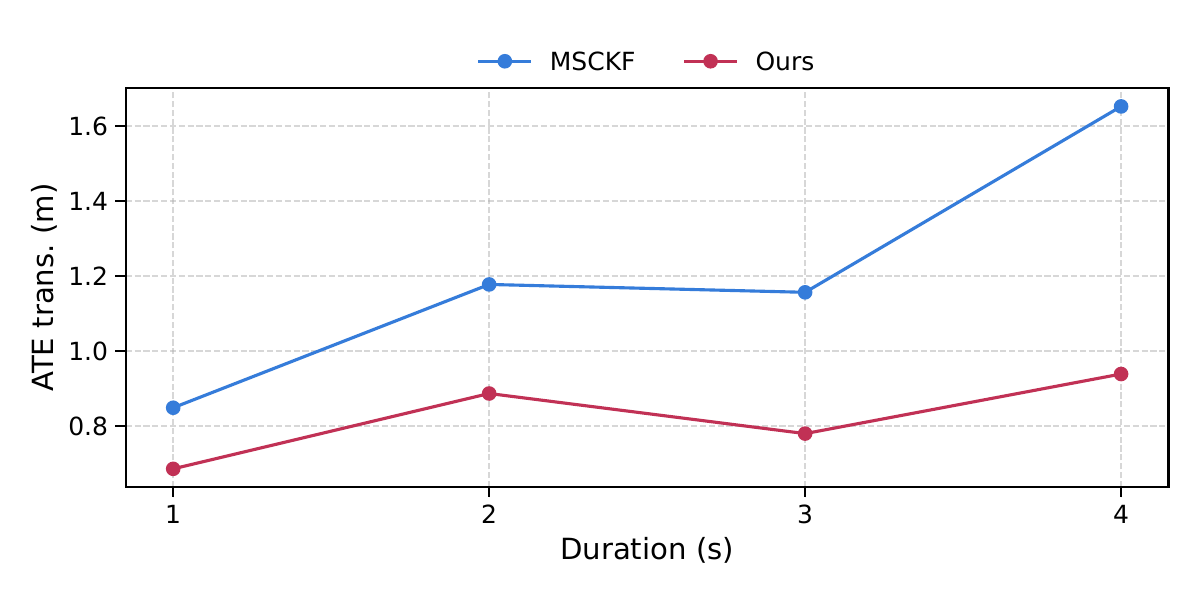}
    \includegraphics[width=1.0\linewidth,trim={0pt 0pt 0pt 10pt},clip]{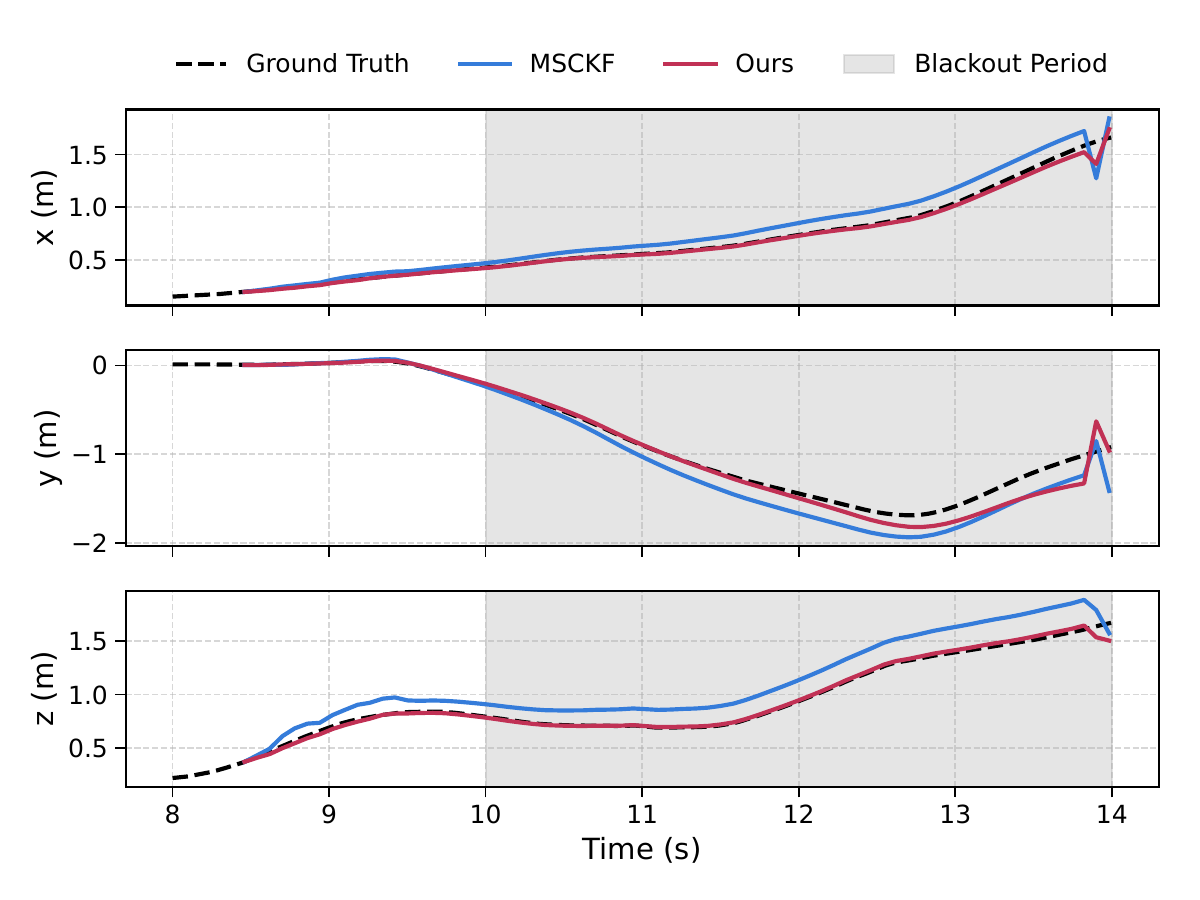}
    \caption{\NEW{(Top) Translational ATE under different visual blackout durations, averaged across \emph{Aerodrome} sequences. (Bottom) Position estimates on \emph{Aerodrome} A03 with a 4-second blackout at 10 seconds.}}
    \label{fig:aerodrome_blackout}
    \vspace{-2mm}
\end{figure}

To demonstrate the benefits of our IMU bias learning approach, we evaluate it in scenarios where visual features are temporarily lost, requiring the filter to rely solely on the IMU measurements for motion estimation. We introduced a single visual feature failure point of durations $1$, $2$, $3$, and $4$ seconds to each of the A03, A04, and A05 sequences of the \emph{Aerodrome} dataset. In Fig. \ref{fig:aerodrome_blackout},  we show the ATE in translation averaged over all three sequences and provide a sample trajectory estimate for the A03 sequence. Our proposed method outperforms the MSCKF, particularly in instances where the filter's bias estimation is inaccurate. The MSCKF estimates bias by minimizing visual measurement residuals, an approach that may not yield the true IMU bias \cite{buchanan2022deep}. In contrast, our method predicts the IMU bias independently of visual information, resulting in improved reliability. Therefore, accurate IMU bias estimation becomes critical during visual feature blackouts to maintain reliable inertial integration. To illustrate this, Fig. \ref{fig:bias_evolution} demonstrates that under normal conditions with consistent visual measurements, MSCKF bias estimates often converge to values already predicted by our method. Additionally, while the true IMU bias typically exhibits slow time-varying behavior, the MSCKF  bias estimates fluctuate, which is inconsistent with expected physical behavior.

\begin{table}[t]
    \caption{Computation times in milliseconds [ms] of our pipeline.}
    \label{tab:computation_time}
    \centering
    \begin{tabular}{cccc}
        \textbf{Processing Step} & \textbf{Min} & \textbf{Mean} & \textbf{Max} \\
        \hline
        \hline
        Tracking & 6.83 & 18.06 & 67.67  \\
        Propagation & 0.28 & 0.77 & 2.48 \\
        Update & 0.03 & 9.19 & 50.01  \\
        {\bf Bias inference} & {\bf 0.56} & {\bf 2.18} & {\bf 9.10}  \\
        Retriangulation and marginalization & 4.35 & 10.11 & 22.47  \\
        Total & 14.99 & 38.14 & 102.62  \\
        \hline
    \end{tabular}
    \vspace{-5mm}
\end{table}

\subsection{Comparison with IMO}
\label{sec:inertial_odometry}
So far, our IMU bias prediction model was used in the MSCKF propagation step. Here, we use it in the update step and \NEW{benchmark against IMO \cite{cioffi2023learned}} for patterned motion without visual information on the Blackbird dataset \cite{antonini2020blackbird}. \NEW{IMO predicts relative positions from world-frame IMU with a neural network and uses them as update measurements for an EKF.} IMO also uses a sliding window, similar to the MSCKF, to attenuate large error instances from the network predictions. 

Our method differs by predicting IMU biases directly from the IMU measurements, correcting the measurements, and integrating to obtain relative position estimates. The biases estimated in this section no longer represent the physical IMU biases alone. Instead, they represent the physical biases \NEW{plus a correction that compensates the unobservable initial velocity so that the integrated position increments align with the ground truth. We compute the relative velocity and position increments $\Delta v_{ij} = v_{j} - v_{i}$ and $ \Delta p_{ij} = p_{j} - p_{i}$ as:}
%
%
{\color{black}
\begin{align}
    \Delta v_{ij} &= \sum_{k=i}^{j-1} \left( R_{k} (\bar{a}_k - \hat{b}^{a}_{k}) + g \right) \Delta t_k,  \\ 
    \Delta p_{ij} &= \sum_{k=i}^{j-1} (v_i + \Delta v_{ik}) \Delta t_k + \frac{1}{2} \left( R_{k} (\bar{a}_k - \hat{b}^{a}_{k}) + g \right) \Delta t_k^2. \nonumber
\end{align}}%
\NEW{Because $\Delta p_{ij}$ depends on the unobservable initial velocity $v_i$, both IMO and our approach implicitly compensate the term $\sum_{k=i}^{j-1} v_i \Delta t_k$. We train only the accelerometer bias with the loss $\| \Delta p_{ij} - \Delta \hat{p}_{ij} \|^2$, where} 
\begin{align}
    \Delta \hat{p}_{ij} &= \sum_{k=i}^{j-1} \Delta v_{ik} \Delta t_k + \frac{1}{2} \left( R_{k} (\bar{a}_k - \hat{b}^{a}_{k}) + g \right) \Delta t_k^2 \nonumber. 
\end{align}
\NEW{Note that $\Delta \hat{p}_{ij}$ omits the initial velocity term. Following \cite{cioffi2023learned}, we use ground-truth $R_k$ during training and the estimated $\hat{R}_k$ from the filter at deployment.} IMO performs well on in-distribution data, achieving an ATE of $0.418$ in translation and $3.028$ in rotation. This accurate performance arises from directly estimating relative positions from rotated IMU measurements in the world frame, simplifying the network’s learning task. However, IMO struggles with out-of-distribution data since it implicitly learns initial velocities specific to patterned motion. In contrast, our method achieves an ATE of $0.772$ in translation and $14.156$ in rotation, yielding relatively accurate estimates along the $x$ and $y$ axes, as presented in Table \ref{tab:blackbird_evaluations}. Our method is less accurate along the $z$ axis due to the double integration of accelerometer measurements $\bar{a}_k$ along with gravitational acceleration $g$ and bias compensation $\hat{b}^{a}_{k}$, which makes the network's learning task more challenging. However, IMO is designed mainly for improving state estimation around known trajectories, e.g., for drone racing. Meanwhile, the primary strength of our approach lies in generalizing to unseen data through IMU bias estimation, which enables the use of an invariant filter in VIO, shown in Sec. \ref{sec:euroc} and \ref{sec:extreme_scenarios}.

\begin{table}[t]
    \caption{\NEW{Inertial odometry on \emph{Blackbird Clover} over 30 seconds.}}
    \label{tab:blackbird_evaluations}
    \centering
    \begin{tabular}{cccccc}
        \textbf{Metrics} & \textbf{Sequence} & \textbf{TLIO \cite{liu2020tlio}} & \textbf{IMO \cite{cioffi2023learned}} & \textbf{Ours} \\
        \hline
        \hline
        ATE x-axis [m] & Clover & $7.340$ & ${\bf 0.357}$ & $\underline{0.431}$  \\
        ATE y-axis [m] & Clover & $3.139$ & ${\bf 0.207}$ & $\underline{0.299}$  \\
        ATE z-axis [m] & Clover & $2.332$ & ${\bf 0.061}$ & $\underline{0.571}$  \\
        \hline
        ATE trans. [m]  & Clover & $8.318$ & ${\bf 0.417}$ & $\underline{0.776}$  \\
        ATE rot. [deg]   & Clover & $75.341$ & ${\bf 3.028}$ & $\underline{7.736}$  \\
        \hline
    \end{tabular}
    \vspace{-5mm}
\end{table}

\section{CONCLUSION}
We developed a learning-based invariant filter for VIO, estimating the IMU bias externally to the filter state via a neural network. This allows us to preserve the system's invariance and achieve robustness over traditional VIO methods, especially in visually degraded scenarios. Future work will focus on learning the measurement uncertainty to assess the reliability of the IMU measurements more accurately and in addition to using additional information as input to the neural network, such as past images or image features.

\bibliographystyle{ieeetr}
\bibliography{bib/ref}

\end{document}